\documentclass[fleqn,11pt]{wlscirep}
\usepackage[parfill]{parskip}
\usepackage{float}
\usepackage{graphicx}
\usepackage{comment}
\usepackage{hanging}

\title{Stovepiping and Malicious Software: A Critical Review of AGI Containment}

\author[1,*]{Jason M. Pittman}
\author[1,+]{Jesus P. Espinoza}
\author[1,+]{Courtney Soboleski Crosby}

\affil[1]{Synthetic Intelligence Research Institute, Capitol Technology University, Laurel MD 20708}
\affil[*]{jmpittman@captechu.edu}
\affil[+]{these authors contributed equally to this work}

\begin{abstract}
Awareness of the possible impacts associated with artificial intelligence has risen in proportion to progress in the field. While there are tremendous benefits to society, many argue that there are just as many, if not more, concerns related to advanced forms of artificial intelligence. Accordingly, research into methods to develop artificial intelligence safely is increasingly important. In this paper, we provide an overview of one such safety paradigm: containment with a critical lens aimed toward  generative adversarial networks and potentially malicious artificial intelligence. Additionally, we illuminate the potential for a developmental blindspot in the stovepiping of containment mechanisms. 
\end{abstract}
 
\begin{document}

\flushbottom
\raggedbottom
\maketitle

\thispagestyle{empty}

\section*{Introduction}

Artificial intelligence (AI) is a timely subject in that it has captured the interest of the general public; businesses have learned to harness AI for marketplace advantage, and researchers are pushing the boundaries of how AI shapes our future. Suffice it to say; modern AI is a realization of perhaps the most essential science abstractions. Nevertheless, the expanding reach of even narrow AI is not without risk or free from concern. Minimally, existing forms of narrow AI are forcing a society-level refactoring (Sotala \& Yampolskiy, 2015). Maximally, more general forms of AI may instantiate unknown or unintended risk. One such form of broad AI is \textit{artificial general intelligence} (AGI). Such intelligence is likely to possess abilities beyond what present and future humans can manifest as well as present a true non-human consciousness (Pfeifer \& Iida, 2004).

There is a growing body of literature exploring the topic of AGI and machine consciousness. To be sure, the idea of superintelligent machines is not new. However, as narrow AI gains prominence through its practical applications, research efforts concerning AGI have decidedly increased as well. Because of the potential risks associated with artificial general intelligence, a robust and prudent subfield has been developed around the safety and trust associated with such systems (Yampolskiy, 2012; Babcock, Kramár, \& Yampolskiy, 2016). Thus, as the field continues to expand, there is a need to collate and critically review the primary research. Doing so enables future work without the need for a time-intensive literature review. Furthermore, corollaries using existing technology (e.g., narrow AI) may likewise benefit future research.

Toward such a goal, Pittman and Soboleski (2018) previously suggested that a measure of continuous integration between intersectional knowledgebases would benefit AGI containment research. Because the overarching problem of how to contain AGI is unresolved, we took that as a general starting point for this critical review of the literature. Further, we adopted the specific critical lens of viewing AGI as potentially malicious software (from the human perspective) (Kott, 2015; Pittman \& Soboleski, 2018) and associated containment propositions as instantiations of \textit{stovepiping}. With such a lens established, we systematically analyzed the literature with the aim of filling in gaps left by the ontological model developed by Pittman and Soboleski.

\section*{Background}

\subsection*{AGI}

Whereas (narrow) AI augments human capability within bounded domains (Stone et al., 2016), AGI is intended to be a sovereign entity with at or beyond human intelligence. The goals of an AGI include further perfection of a set of primal functions; chiefly, "self-preservation, self-improvements, and resource acquisition" (Babcock, Kramár, \& Yampolskiy, 2017, pp. 1-2). However, Babcock et al. (2016) reaffirmed that humanity is many decades behind fully developed AGI. Still, the growing desire for better and smarter AI motivates the development of AGI (Nasirian, Ahmadian, \& Lee, 2017).

Along with such lines, Babcock et al. (2016) argued that the first AGI could evolve from a collection of narrow AI. For instance, Babcock et al. described an AGI that evolved from an optimized AI that plays Super Mario Brothers. However, others suggested that aggregation of narrow AI will not prove a viable mechanism to develop AGI (Baillie, 2016). Regardless of \textit{if} AGI is reachable, there is consensus across the literature regarding a need to codevelop safety and trust mechanisms as a consequence of having decided to pursue AGI. Such mechanisms stem from suggestions that the first AGI could act on evasion techniques that deceive natural intelligence and underestimate the perceived benefits and characteristics of an intelligence beyond human (Babcock et al., 2017). Accordingly, there is a growing body of scientific inquiry into means to pursue machine superintelligence with little or no risk (Babcock et al., 2017).

\subsection*{A General AGI Containment Overview}

Containment is a logical consequent of the notion that AGI could act discordant to human interest. Babcock et al. (2016), Sotala and Yampolskiy (2015), and Babcock et al. (2017) suggested that AGI ought to exist within a specific safety construct referred to as \textit{containment} or a \textit{box}. Functionally, containment or boxing is intended to prevent, detect, or otherwise control the ingress and egress of interactions between an AGI and humans. Conceptually though, both artificial general intelligence and the containment of such entities are demanding abstractions. This notion is not surprising given that the nature of autonomy itself is hidden from our understanding and instrumentation (Wiese, 2018).

The language to describe such abstractions are underdeveloped (Pittman \& Soboleski, 2018), and not yet of a fidelity sufficient to pursue a practical application. However, there are analogs to which we can turn to gain approximate understanding. According to Babcock et al. (2017), the domain of AI safety includes such fields as image recognition and language recognition, for example. Even within AI safety domains, there exists adversarial AI (generative adversarial network or GAN), which may attack another AI as demonstrated in the studies of Bose and Aarabi (2018), Carlini et al. (2016), and Song, Shu, Kushman, and Ermon (2018). Furthermore, adversarial AI for cybersecurity penetration-testing is another aspect of AI safety (Masuya \& Takaesu, 2018) when converged with Cyber Science.

\subsection*{Illustrating Containment Through Adversarial Space}

We present three examples of existing AI safety domains to approximate ambiguity in our understanding of future AGI containment. First, voice assistant systems (VAS) reason and integrate various elements of narrow AI, such as the ability to track events, people, and locations (Paul, Bhat, \& Lone, 2017). Positively, Paul et al. (2017) described the functionality of VAS such as \textit{Cortana}, \textit{Siri}, and \textit{Google Assistant} as beneficial. Further, the field survey of Nasirian et al. (2017) on a plurality of VAS determined that the performance quality of such systems influence user trust and adoption of narrow AI, where performance and confidence are linear. However, such systems also incite concerns about the security implications of massive VAS adoption due to attacks that subvert machine understanding (Carlini et al., 2016). That is, the executed hidden voice commands described by Carlini et al. (2016) were unintelligible to humans while specific surreptitious commands were understood and executed by the VAS.

Furthermore, the physical attributes of a hidden voice command attack include matter that is non-organic. Thus, the visibility of the attack may be in the audible spectrum or otherwise in the infrasonic or ultrasonic range. Also, the locality depends on the range of the unintelligible audio and location of the input sensor (e.g., Microphone) (Carlini et al., 2016), particularly where the architecture of such an attack is composed of a combination of hardware and software. In such a scenario, hidden voice attacks have the goal to evade AI detection and confuse natural intelligence. In the specific experimental scenario carried out by Carlini et al. (2016), the attacker knew the inner workings of the vulnerable AI and used logic targeting a white-box model. Conversely, non-equilibrium was a consideration when the offense was carried against a black-box model (Carlini et al., 2016).

Similar to hidden voice attacks, AI for image classification presents vulnerabilities to covert channels when adversarial attacks subvert machine learning neural networks in a manner that is unintelligible to natural intelligence (Song et al., 2018). Specifically, generative adversarial attacks test the security AI and deceive natural intelligence using a training model called auxiliary classifier generative adversarial network (AC-GAN) (Song et al., 2018). Used antagonistically, the experiments of Song et al. (2018) show that GANs overcome the problem presented by Carlini et al. (2016) of generating purposely misclassified images while simultaneously evading traditional adversarial defenses. In short, evasion attacks against machine learning classifiers are an active attack initiated by human intelligence further reasoned by generative adversarial attacks that acquire knowledge and skills found in non-malicious AI (Carlini et al., 2016).

In our third example, the relationships occurring in the cyber world during attacks by weaponized AI are areas of concern for cyberscience AGI containment (Pittman \& Soboleski, 2018). The fundamental requirements of a cyber attack against an agent exist when an adversarial agent violates a security policy to exploit vulnerable AI (Pittman \& Soboleski, 2018). For example, an adversarial agent disturbing a vulnerable agent by executing responsive and preemptive attacks will generate environmental states of non-equilibrium (Goodfellow et al., 2014; Pittman \& Soboleski, 2018).

The experiments conducted by Masuya and Takaesu (2018) led to the development of a penetration-testing tool based on deep learning and string pattern matching called \textit{GyoiThon} as opposed to the simple coupling of penetration- testing stages. First, GyoiThon crawled a URL to obtain as many HTTP responses to gather sufficient data for learning and improved accuracy (Masuya \& Takaesu, 2018). The targets were identified through a learning algorithm using Naive Bayes to determine the software and version of the HTTP service (e.g., Apache, HeartCore, Joomla) (Masuya \& Takaesu, 2018). Next, \textit{GyoiThon} was communicatively coupled to Metasploit, a complete penetration testing framework, through remote procedure calls and an application programming interface (Masuya \& Takaesu, 2018). Afterward, \textit{GyoiThon} controlled Metasploit to conduct the exploitation of vulnerabilities that targeted a specific cloud service, software type, and version (Masuya \& Takaesu, 2018). Indeed, the novel experiments of Masuya and Takaesu (2018) are a step forward into narrow AI for penetration-testing. Effectively, the study described by Masuya and Takaesu (2018) caused the natural intelligence to become a passive agent during reasoning and execution of penetration-testing audits, which is a more rigorous endeavor for the safeguarding of information technologies.

\subsection*{Malicious software}

Now, such containment illustrations become heightened when artificial intelligence becomes autonomously malicious through its own accord as opposed to intentionally weaponized. Malicious software is a behavioral abstraction applied to runtime instantiations of software. While the kind of maliciousness that could be potentiated in AGI, as described by Bostrom (2014) for instance, is possible, there exist more probable malicious behaviors in narrow AI that may yield insight into proper containment modalities. This review took guidance from Baudrillard (1994) in that an artificial reality becomes better than a fundamental reality when lives are predominantly artificial. Furthermore, the philosophy of GANs exhibit internally consistent (or closed) malicious behavior. As an example containment strategy, GAN is meaningful as such embody the notion of using narrow AI to safeguard AGI (e.g., narrow AI as a gatekeeper) and exist as a technology today.

Thus, the practical application of such assumptions is observable in the intersection of postmodern philosophy and the game theory of a generative adversarial attack (Baudrillard, 1994, p. 3).  Specifically, Baudrillard described a society at the edge of media overload where the meaning of images is infinitely mutable. Mutability is at the very essence of GANs where a discriminator \textit{D} and a generator \textit{G} collapse into a state of equilibrium or equality (Goodfellow et al., 2014). Goodfellow et al. (2014) described a GAN as a system with multiple perceptrons fighting against each other for learning and inference. A GAN reached a unique solution when a Generator \textit{G} solved an intractable probabilistic problem, and a discriminator \textit{D} converged to a probability of 0.5 for binary classification (Goodfellow et al., 2014).

First, the discriminator \textit{D} would train for half the game with images adequately classified and under supervision (Goodfellow et al., 2014).  At the beginning of the game, the image set could be considered as “the reflection of a profound reality” (Baudrillard, 1994, p. 6) and of good appearance. Second, the generator \textit{G} employs random vectors to exploit the weaknesses of the discriminator D while both become increasingly better at playing a minimax game (Goodfellow et al.,  2014).  The intent of the images from the Generator \textit{G} is to deceive the discriminator \textit{D} (Goodfellow et al., 2014). The meaning of this stage is to produce images that “mask and pervert a profound reality” (Baudrillard, 1994, p. 6) with evil appearance. In the third stage, the generator \textit{G} is successful at fooling the discriminator with infinitely mutable images that are always to be believed to be the truth (Goodfellow et al., 2014). Baudrillard (1994) would refer to such capability as “the absence of a profound reality” (p. 6) while maintaining appearance. Finally, game theory would expect a well-advanced generator \textit{G} to reach a point of equilibrium where the images could become better than real or hyperreal (Baudrillard, 1994; Goodfellow et al., 2014). The teachings of Goodfellow et al. (2014) and Baudrillard (1994) emphasized the philosophical and computational requirements of \textit{maleficence} to achieve equilibrium in a simulation.

\subsection*{Stovepiping}

Commonly, the equilibrium control mechanism applied to malicious runtime software is \textit{antimalware} software. That is, a form of software operating as an overwatch or governor. While traditional antimalware is not a viable containment mechanism itself, the concept may hint at a solution. For example, the use of a narrow AI as an overwatch or governor type mechanism for AGI containment is not without precedence (Babcock et al., 2017). Moreover, in addition to the plethora of existing viable technologies, the likelihood of developing containment purpose-specific apparatuses in the future is high. However, classical approaches (i.e., antivirus) are notoriously \text{stovepiped} (Sun, Dai, Singhal, \& Liu, 2017).

Thus, a problem may exist insofar as containment constructs are not vertically or horizontally integrated (Babcock et al., 2016; Babcock et al., 2017). The lack of dimensional integration is referred to as stovepiping (Boehm, 2005). Stovepiping promotes risk concentration and inhibits information sharing. Thus, while the result of technologies that function in a binary blacklist-whitelist manner may demonstrate limited control, stovepipes run counter to the very nature of containment by enforcing a strict non-equilibrium across the containment environment. The inherent flaw is that such non-equilibrium works only to the extent that equilibrium is not restored. As an AGI experiences, that experience will alter reality for the AGI. In doing so, an AGI may restore equilibrium across the containment environment inadvertently. Just as likely, the AGI may perceive the non-equilibrium and seek to restore such through direct machinations. The result of either scenario would undermine the very intent of containment as long as the mechanism(s) of containment and array of specific controls are not unified.

As well, a lack of dimensional integration may allow for pockets of non-control throughout any form of containment environment. Indeed, according to Boehm (2005), stovepiping ultimately results in zones of isolated vulnerability. Such vulnerabilities are exacerbated by complex or dynamic operating environments. We suggest AGI- moreover, AGI in containment- represents a dynamic (perhaps hostile) operating environment. Further, AGI and related containment features represent undeniably complex computing architectures. Ultimately, stovepiping may lead to AGI leakage across the containment barrier.

\section*{Conclusions}

AGI research is germane given the influx of advancements in narrow AI. Thus, it is prudent to investigate concerns around mechanisms to ensure AI safety. Necessarily, AI safety induces discussions of how to contain or box potentially malicious AGI given that AGI is an inherently and experientially dynamic entity. This critical review described a developmental blindspot in containment constructs that may be stovepiped during control routines at the expense of dynamism (Boehm, 2005). Specifically, the examples including VAS, image classification, and cybersecurity penetration-testing demonstrated the applications of malicious AI (Carlini et al., 2016; Masuya \& Takaesu, 2018; Song et al., 2018).

Furthermore, an operationally malicious AGI would follow philosophical and computational methods that create a better reality or hyperreality with the intent of achieving an equilibrium through exploitable elements in the containment environment. The inputs of an AGI may be random due to the system and method of adversarial machine learning (Goodfellow et al., 2014). In this simulated state, the AGI may have achieved a positive or negative (im)balance within the containment environment due to uncertainty while maintaining appropriate outputs for compliance (Baudrillard, 1994; Goodfellow et al., 2014).

While AGI containment is not a solved problem, existing paradigms in the adversarial space can help disambiguate the solution set. Adversarial narrow AI, such as GANs, demonstrates a profound understanding of the underlying containment mechanisms and instrumentation towards the use of more general applications. Henceforward, the problem of designing containment barriers that are resistant to deception is of importance to safety engineering and germinal research of adversarial machine learning for viable AGI containment.

\medskip

\section*{References}
\begin{hangparas}{.25in}{1}
\setlength{\parskip}{0em} 
Babcock, J., Kramár, J., \& Yampolskiy, R. (2016). The AGI containment problem. \textit{Cornell University Library, abs/1604.00545}. arXiv: 1604.00545. Retrieved from http://arxiv.org/abs/1604.00545

Babcock, J., Kramár, J., \& Yampolskiy, R. V. (2017). Guidelines for artificial intelligence containment. \textit{Cornell University Library, abs/1707.08476}. arXiv: 1707.08476. Retrieved from http://arxiv.org/abs/1707.08476

Baillie, J.-C. (2016). Why alphago is not AI. \textit{IEEE spectrum posted, 17.} Retrieved from https://spectrum.ieee.org/
automaton/robotics/artificial-intelligence/why-alphago-is-not-ai

Baudrillard, J. (1994). \textit{Simulacra and simulation}. Michigan, USA: The University of Michigan. Retrieved from https://books.google.com/books?id=9Z9biHaoLZIC

Boehm, B. (2005). The future of software processes. \textit{Proceedings of the 2005 international conference on unifying the software process spectrum,} Beijing, China: Springer-Verlag, 10–24. doi:10.1007/11608035\_2

Bose, A. J. \& Aarabi, P. (2018, May). Adversarial attacks on face detectors using neural net based constrained optimization. \textit{Cornell University Library}. Retrieved from https://arxiv.org/pdf/1805.12302.pdf

Bostrom, N. (2014). \textit{Superintelligence: Paths, dangers, strategies} (1st). New York, NY, USA: Oxford University Press, Inc.

Carlini, N., Mishra, P., Vaidya, T., Zhang, Y., Sherr, M., Shields, C., . . . Zhou, W. (2016). Hidden voice commands. \textit{25th usenix security symposium (usenix security 16)}, Austin, TX. Retrieved from https://nicholas.carlini.com/
papers/2016\_usenix\_hiddenvoicecommands.pdf

Goodfellow, I., Pouget-Abadie, J., Mirza, M., Xu, B., Warde-Farley, D., Ozair, S., . . . Bengio, Y. (2014). \textit{Generative adversarial nets}. Curran Associates, Inc. Retrieved from http://papers.nips.cc/paper/5423-generative-adversarial-nets.pdf

Kott, A. (2015). Science of cyber security as a system of models and problems. \textit{CoRR, abs/1512.00407}. arXiv: 1512.00407. Retrieved from http://arxiv.org/abs/1512.00407

Masuya, M. \& Takaesu, I. (2018). Gyoithon. Retrieved from https://github.com/gyoisamurai/GyoiThon

Nasirian, F., Ahmadian, M., \& Lee, O. D. (2017, May). AI-based voice assistant systems: Evaluating from the interaction and trust perspectives. \textit{Twenty-third Americas conference on information systems}, Boston, US, 1–10. Retrieved from https://www.researchgate.net/publication/322665841\_AI-Based\_Voice\_Assistant\_
Systems\_Evaluating\_from\_the\_Interaction\_and\_Trust\_Perspectives

Paul, Z. M., Bhat, H. R., \& Lone, T. A. (2017, August). Cortana-intelligent personal digital assistant: A review. \textit{International Journal of Advanced Research in Computer Science, 8}, 55–57. doi:10.26483/ijarcs.v8i7.4225 

Pfeifer, R. \& Iida, F. (2004). \textit{Embodied artificial intelligence: Trends and challenges}. Springer. doi:10.1007/978-3-540-27833-7\_1

Pittman, J. M. \& Soboleski, C. E. (2018). A cyber science based ontology for artificial general intelligence containment. \textit{Cornell University Library, abs/1801.09317}. arXiv: 1801.09317. Retrieved from http://arxiv.org/ abs/1801.09317

Song, Y., Shu, R., Kushman, N., \& Ermon, S. (2018). Generative adversarial examples. \textit{Cornell University Library}. arXiv: 1805.07894. Retrieved from https://arxiv.org/pdf/1805.07894.pdf

Sotala, K. \& Yampolskiy, R. V. (2015). Responses to catastrophic AGI risk: A survey. \textit{Physica Scripta, 90}(018001), 1--33. doi:10.1088/0031-8949/90/1/018001 

Stone, P., Brooks, R., Brynjolfsson, E., Calo, R., Etzioni, O., Hager, G., . . . Kraus, S., et al. (2016). Artificial intelligence and life in 2030. One hundred year study on artificial intelligence: Report of the 2015-2016 study panel. \textit{Stanford, CA: Stanford University, 7}, 1--52. Retrieved from http://ai100.stanford.edu/2016-report

Sun, X., Dai, J., Singhal, A., \& Liu, P. (2017). Enterprise-level cyber situation awareness. \textit{Theory and models for cyber situation awareness,} Springer, \textit{10030,} 66--109. doi:10.1007/978-3-319-61152-5\_4

Wiese, W. (2018). Toward a mature science of consciousness. \textit{Frontiers in Psychology, 9}, 693. doi:10.3389/fpsyg.
2018.00693

Yampolskiy, R. V. (2012). Leakproofing singularity-artificial intelligence confinement problem. \textit{Journal of Con-sciousness Studies JCS, 19}(1-2), 194--214. Retrieved from http://cecs.louisville.edu/ry/LeakproofingtheSingularity. pdf

\end{hangparas}

\end{document}